\documentclass{article}

\usepackage{microtype}
\usepackage{graphicx}
\usepackage{subfigure}
\usepackage{booktabs} 
\usepackage{apacite}
\usepackage{amssymb}
\usepackage{amsmath}

\usepackage{hyperref}

\usepackage[accepted]{icml2020}

\icmltitlerunning{Under the Hood of Neural Networks}

\begin{document}

\twocolumn[
\icmltitle{Under the Hood of Neural Networks: Characterizing Learned Representations by Functional Neuron Populations and Network Ablations}



\icmlsetsymbol{equal}{*}

\begin{icmlauthorlist}
\icmlauthor{Richard Meyes}{buw}
\icmlauthor{Constantin Waubert de Puiseau}{buw}
\icmlauthor{Andres Posada-Moreno}{rwth}
\icmlauthor{Tobias Meisen}{buw}
\end{icmlauthorlist}

\icmlaffiliation{buw}{Institute of Technologies and Management of Digital Transformation, Wuppertal, Germany}
\icmlaffiliation{rwth}{Institute of Information Management in Mechanical Engineering, Aachen, Germany}

\icmlcorrespondingauthor{Richard Meyes}{meyes@uni-wuppertal.de}

\icmlkeywords{Transparency, Learned Representations, Neuroscience, Neuron Populations, Network Ablations}

\vskip 0.3in
]



\printAffiliationsAndNotice{\icmlEqualContribution} 

\begin{abstract}
The need for more transparency of the decision-making processes in artificial neural networks steadily increases driven by their applications in safety critical and ethically challenging domains such as autonomous driving or medical diagnostics. We address today's lack of transparency of neural networks and shed light on the roles of single neurons and groups of neurons within the network fulfilling a learned task. Inspired by research in the field of neuroscience, we characterize the learned representations by activation patterns and network ablations, revealing functional neuron populations that a) act jointly in response to specific stimuli or b) have similar impact on the network's performance after being ablated. We find that neither a neuron's magnitude or selectivity of activation, nor its impact on network performance are sufficient stand-alone indicators for its importance for the overall task. We argue that such indicators are essential for future advances in transfer learning and modern neuroscience.
\end{abstract}

\section{Introduction}
Research on deep learning has brought forth a number of remarkable applications in the domains of computer vision, natural language processing, self-learning agents and continuous control covering the fields of artificial vision, speech and motion. The main research focus in the past was placed on increasing the size, performance and speed of deep neural networks solving specific benchmark tasks or refining the training algorithms to tackle increasingly complex problems. In some cases, this has led to unexpected and/or unwanted behavioral artifacts of trained networks \cite{yudkowsky2008artificial,  ribeiro2016should, winkler2019association}. Recently, efforts to explain the occurrence of such artifacts were made with the development of methods to investigate the organization and structure of the learned representations in these increasingly complex networks. Such methods form the fundamental basis in the field of neuroscience, where large and complex neural systems have been the objects of investigation for decades, aiming to make such systems transparent and explainable. 

In this paper, we follow a neuroscience inspired approach to investigate the learned representations of a neural network based on its activity in response to specific inputs, an approach that has been used for over half a century to understand functions of the mammalian brain \cite{hubel1959receptive} and based on its robustness against partial network ablations. Specifically, we trained three instances of a network on three different image classification datasets and observed how the learned representations evolve along the hierarchically structured layers of the network and how these representations are affected by partial ablations of the network. We further investigate the networks for clusters of functional neuron populations consisting of jointly operating neurons. 

We found that the representations evolve to become more distinct, effectively improving the separation of the classes in the activation-space of the network towards the output layer. We further found that ablations impact the separation in the activation-space leading to an overlap of classes and thus, false classifications of inputs. We further found distinct activity patterns in the network's activation-space for the different classes. Specifically, similar to findings in the mammalian brain \cite{nakamura1998somatosensory, kaschube2008self, da2011human}, a distinct set of units shows a high activity in response to specific inputs and a low activity for other inputs. Furthermore, visualizing single unit activity according to their selectivity for specific inputs revealed a strong variation in the amount of units that are most selective for the different classes. Additionally, comparing the effects of ablations on the class specific classification accuracy of the network revealed that the importance of a single unit for the classification task cannot solely be attributed to the magnitude or the selectivity of its activity. 

\section{Related Work}
Recent efforts addressed the increased demand for transparency of AI-driven applications in domains such as production technology, medical diagnostics or autonomous driving, where mistakes can have potentially devastating consequences from an economic, ethic and legal point of view, and have spurred research in the field of explainable and interpretable AI \cite{lipton2016mythos, goodman2017european, su2019one}. A main goal was placed on investigating dependencies of a network's output on its input. The most prominent methods for explaining the influences of input variables on the prediction of a network include methods like gradient based class activation mapping (Grad-CAM) \cite{selvaraju2017grad}, layer-wise relevance propagation (LRP) \cite{binder2016layer}, deep taylor decomposition \cite{montavon2016deep} and class-enhanced attentive response (CLEAR) \cite{kumar2017explaining}, which allow to determine what specific input features contribute to the decision of a neural network. These dependencies facilitated a number of perturbation studies, in which input images were manipulated systematically, showing that only marginal modifications and even single pixel alterations can drastically change the prediction of the network \cite{papernot2017practical, fong2017interpretable, faust2018visualizing, su2019one}. These studies, however, merely focus on the processed data and the result of a neural network and disregard their inner processes and learned representations, when explaining their decision making processes.

Aiming to look inside of neural networks, a number of tools have been developed to visualize network activity in response to specific inputs \cite{harley2015interactive} or in response to varying hyper parameters \cite{smilkov2017direct} as well as to compare models with each other \cite{zhang2018manifold}. These tools visualize learned filters and feature maps of CNNs \cite{chung2016revacnn}, allowing to investigate them for their similarities and representative power for specific classes \cite{liu2016towards} or specific input features \cite{olah2017feature, olah2018building}. Furthermore, single units have been identified that only contribute negligibly to a task and can be pruned \cite{molchanov2016pruning}, that align with with semantic concepts in images \cite{bau2017network} or represent linguistic properties or sentiment in texts \cite{bau2018identifying, radford2017learning}. Recently, Activation Atlases of whole large scale networks extended the mere visualization of features by giving spatial meaning to them via two-dimensional embeddings of the networks' activations \cite{carter2019activation}, confirming previous findings on the location of global features and class specific features within the network \cite{zeiler2014visualizing}. Embedding methods such as t-SNE \cite{maaten2008visualizing} or UMAP \cite{mcinnes2018umap} have been widely used to visualize the high dimensional activation-space of neural networks to identify the role of network areas in solving a given task \cite{liu2016towards, rauber2016visualizing, elloumi2018analyzing, dibia2019convnet}. 

Alternatively to investigating network activations, network ablations have been used to study the effect of single units on a network's performance \cite{dalvi2019neurox}, helping to decided what units can be pruned with minimal effect on a network's discriminative power \cite{li2016pruning, cheney2017robustness}. Targeted ablations in GANs trained to generate photorealistic images were used to delete specific objects such as chairs or windows from the generated images \cite{bau2019visualizing}. Determining what makes a single unit important for solving a task, it has been shown that ablating units with large weights has a stronger impact on network performance than ablating units with small weights \cite{dalvi2019one}. Complementary, it has been shown that the importance of a unit is not only determined by the magnitude of it's weights, but rather by the extent to which the distribution of it's incoming weights changes during training \cite{meyes2019ablation, meyesablation}. Additionally taking a unit's activation into account, it has been shown that units with a high class selectivity, which are easily interpretable, are not necessarily more important for the overall task than units with a low class selectivity and a less accessible interpretability \cite{morcos2018importance}. Recently, controversial insights on methods how to evaluate the similarity of learned network representations have been reported and demonstrate the early stage of current knowledge and thus, the importance and the need for more research on the topic \cite{morcos2018insights, kornblith2019similarity}. 

We complement the related work by a combined approach of investigating embeddings of network activations of healthy and ablated networks revealing functional neuron populations with distinguishable significance for the learned representations. 

\section{Methods}
\subsection{Network Training and Ablations} \label{ssec:ann_setup_and_abl}
We trained a custom network that consists of three convolutional layers, \textit{"conv1"}, \textit{"conv2"} and \textit{"conv3"}, two fully connected layers, \textit{"fc1"} and \textit{"fc2"} and the output layer \textit{"out"}. All convolutional layers feature $64$ 2-D kernels of size $5\times5$ with a stride of $1$ and zero-padding of $2$ and are followed by max-pooling layers with $2\times2$ kernels and stride $2$. The fully connected layers are comprised of $512$ neurons each while the output layer is comprised of $10$ neurons. ReLU activation is chosen for all layers except the output layer, which uses log-softmax activation. Separate instances of the network were trained on the normalized $(\mathcal{N}(\mu=0.5,\,\sigma=0.5))$ MNIST \cite{lecun-mnisthandwrittendigit-2010}, Kuzushiji-MNIST \cite{clanuwat2018deep} and Fashion-MNIST \cite{xiao2017fashion} dataset for $100$ epochs with a learning rate of $0.001$ and momentum of $0.9$, optimizing the cross-entropy loss with stochastic gradient descent for the ten target classes. The $60,000$ training images per dataset were processed with a batch size of $64$. Testing was conducted using $9984$ out of $10,000$ test images due to a test batch-size of $32$. Henceforth, the three networks will be referred to as \textit{"M-Net"}, \textit{"K-Net"} and \textit{"F-Net"}. All networks were implemented and trained with PyTorch v1.3 \cite{paszke2019pytorch} and scored top-1 accuracies of $99.0\%$, $95.3\%$ and $91.2\%$, on the MNIST, KMNIST and Fashion-MNIST dataset, respectively. All computations were performed on a single end consumer machine containing an 8 core Ryzen 7 1800x processor and a single NVIDIA GTX 1080Ti GPU.

Ablations of single neurons in the fully connected layers were performed by manually setting their incoming weights and biases to zero, effectively preventing any flow of information through those neurons. Concurrently, ablations in convolutional layers were performed by setting the weights and biases of all neurons of a kernel to zero, consequently ablating $5\times5$ neurons at once. For reasons of simplicity, ablated single units in the fully connected layers as well as ablated kernels are referred to as units throughout the remainder of this paper. 

\subsection{Embedding of Network Activations} \label{ssec:inv_actspace}
\label{sec:embedding_of_net_act}
Activations of each unit of the three networks in response to each image in the three test sets were stored in a matrix. Considering the number of test images ($9,984$) and the number of neurons in each network ($17,280$), the resulting activation matrices are $M_{X-Net} \in \mathbb{R}^{9984\times17280}$, where $X \in \{M, K, F\}$. The activation matrices were embedded using UMAP in two ways. Either dimension of the matrix was reduced, so that a point either represents the activation of the whole network or a single network layer in response to a single test image (horizontal reduction, $M \in \mathbb{R}^{9984\times2}$) or it represents the activation of a single unit in response to the whole test set (vertical reduction,  $M \in \mathbb{R}^{2\times17280}$). We used an open source Python implementation of UMAP \cite{mcinnes2018umap} with default parameters after an initial attempt for finding better values for the number of nearest neighbours or the minimum distance between data points yielded no significant visual improvement of the embeddings. 

\begin{figure}[tb!]
\vskip 0.2in
\begin{center}
\centerline{\includegraphics[width=0.48\textwidth]{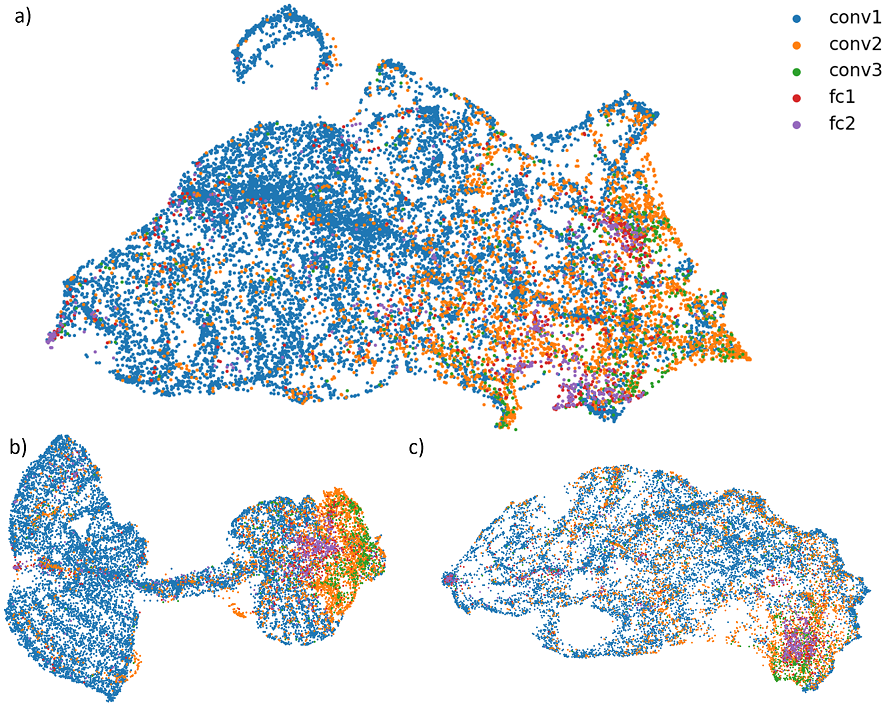}}
\caption{Neuron populations obtained from the vertical reduction of the activation-space. a) \textit{F-Net}, b) \textit{K-Net} and c) \textit{M-Net}.}
\label{fig:Neuron populations overview}
\end{center}
\vskip -0.2in
\end{figure}

In order to make the activation embeddings after horizontal reduction of different network layers comparable to each other, embeddings were initialized by applying UMAP directly to the test set with loosened constraints ($min\_dist = 0.8$) so that the initial coordinates of the data points for each embedded layer activation were the same. Since linear shifts, scales and rotations are not accounted for by UMAP, we used Scipy's Procrustes transformation \cite{Scipy2019arXiv} to linearly scale, shift, reflect and rotate the embeddings with respect to the projection of the previous layer, which further improved the comparability between activation embeddings. We used the neighborhood hit (NH) in the activation embeddings as a quantitative measure of class separation. The NH-score is a measure for the percentage of points, for which the k nearest neighbors of a point belong to the same class as the point itself. We empirically determined $k = 6$ to yield reasonable results which are consistent with our visual inspections. Aiming to investigate network activity for functional neuron populations, i.e. clusters of neurons with similar activations in response to the test images, we assigned different colors to each unit in the vertically reduced embeddings. Figure \ref{fig:Neuron populations overview} shows the three neuron populations of \textit{M-Net}, \textit{K-Net} and \textit{F-Net}, with each unit being colored according to its layer affiliation. 

We alternatively colored the neurons according to functional metrics characterizing a neuron's magnitude of activation or its class selectivity (not shown in Figure \ref{fig:Neuron populations overview}). As a measure for how selectively a neuron activates for a specific class, the activational selectivity (AS), defined as
\begin{equation*}
AS = \frac{\mu_{max} - \mu_{av. else}}{\mu_{max} + \mu_{av. else}} \in [0, 1]
\end{equation*}
was calculated for each neuron, where \(\mu_{max}\) is the highest class-specific mean activity and \(\mu_{av. else}\) is the mean activity across all other classes \cite{morcos2018importance}. Higher values of AS denote a stronger tendency of a unit to only activate for a single class. In cases, in which the denominator was $0$, we manually set the AS to $0$. As a measure of characterizing a units importance for representing a class, we colored the neurons based on the change of network accuracy as a result of the ablation of that neuron. Similarly to the AS, the ablation effect selectivity (AES), is defined as
\begin{equation*}
AES = \frac {\Delta_{max} - \Delta_{av. else}} {\Delta_{max} + \Delta_{av. else}} 
\end{equation*}
where \(\Delta_{max}\) is the highest class-specific change in accuracy and \(\Delta_{av. else}\) is the average change in accuracy across the other classes. Since the AES can be positive or negative, we separated the scale into two and re-scaled both according to their maximum positive or negative values to take values between $0$ and $1$. Analogously to the AS, in cases, in which the denominator was $0$, the AES was set to $0$.

\section{Results}
\subsection{Network Ablations}
We performed network ablations in different layers to determine whether the representation of the different classes is equally distributed across the network or whether it shows a preference for some layers over the others. Initially performing ablations of $10\%$, $20\%$, $30\%$, $40\%$ and $50\%$ of units within a single layer, we found that all three networks are fairly robust against ablations, showing only marginal changes in accuracy for smaller amounts. Thus, for the remainder of this paper, we report results on ablations of $50\%$ of units within a layer. For the ablations, we performed a random selection without replacement of units to be ablated $100$ times and calculated the average change in accuracy for the specific classes. 

\begin{figure}[b!]
\vskip 0.2in
\begin{center}
\centerline{\includegraphics[width=0.48\textwidth]{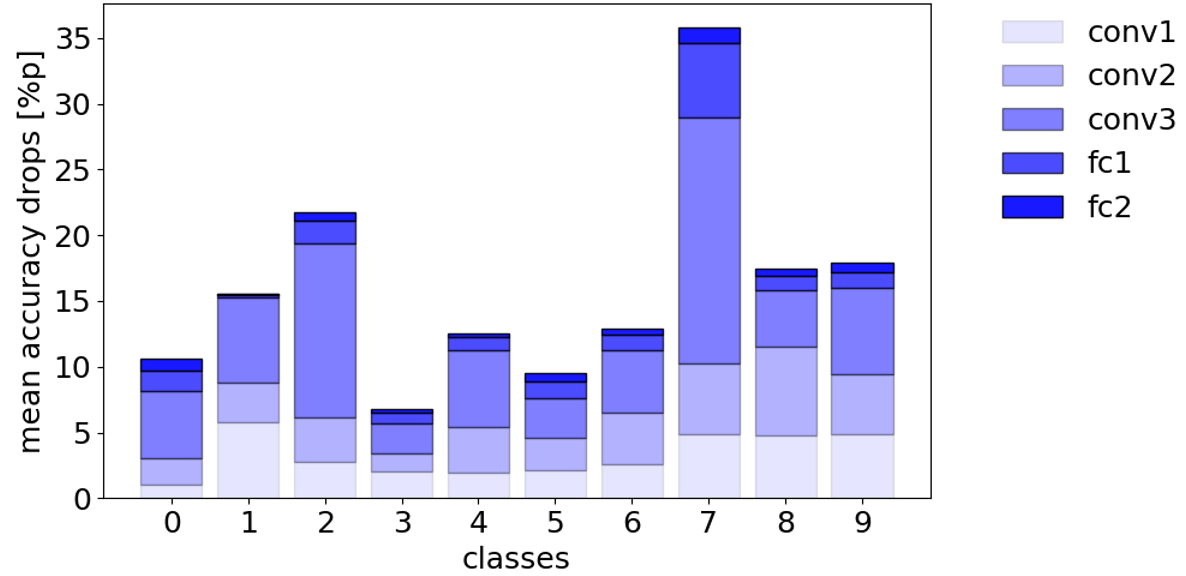}}
\caption{Stacked bar plot of mean accuracy changes as a result of network ablations in \textit{M-Net}.}
\label{fig:stacked_barplot_Mnet}
\end{center}
\vskip -0.2in
\end{figure}

Figure \ref{fig:stacked_barplot_Mnet} shows the effects of ablations in the \textit{M-Net}.
Some classes are more severely affected by the ablations than other classes, suggesting that the amount of capacity used to represent the classes differs greatly. For instance, class 7 shows the largest change in accuracy while class 3 shows the smallest, indicating that the network used much more of its capacity to represent class 7 compared to class 3. This implies that the representation of class 7 is more complex than the representation of class 3. However, this is not directly explainable based on the mere separability of the classes. More precisely, although class 3 and class 7 are equally well separated in the embedded data space (cf. Figure A3 in the Appendix), the amount of network capacity required for their representations differs greatly. Another explanation may be that class 3 is represented more redundantly than class 7, so that ablations have a less severe effect. Figure \ref{fig:stacked_barplot_Mnet} further shows that the representations of single classes is distributed across the layers. For instance, the distribution of the classes 2 and 7 show the strongest localization in \textit{conv3} compared to the other layers, which is consistent with the notion that the last convolutional layer functions as the feature extraction layer, which represents the most distinct features of the data. However, the classes 1 and 8 and 9 for example show a more equal distribution across the three convolutional layers.

We tested whether the difficulty to predict a class correlates with the amount of capacity that is reserved by the network to represent that specific class. To this end, we calculated the Spearman rank-order correlation between the original prediction error and the change in accuracy for each class. A correlation coefficient of $r=0.6$ and a p-value of $p=0.07$ for \textit{K-Net} and $r=0.48$, $p=0.16$ for \textit{F-Net} suggests, that classes, which are more difficult to predict are also more sensitive to ablations. The corresponding bar plots can be found in Figures A1 and A2 in the Appendix. Note, however, that the number of samples of $10$ is small limiting the descriptive statistical power of the test. In case of \textit{M-Net} (cf. Figure \ref{fig:stacked_barplot_Mnet}), no significant correlation was found ($r=0.26$, $p=0.47$) 

\subsection{Evolvement of Representations}
\begin{figure}[htb!]
\vskip 0.2in
\begin{center}
\centerline{\includegraphics[width=0.48\textwidth]{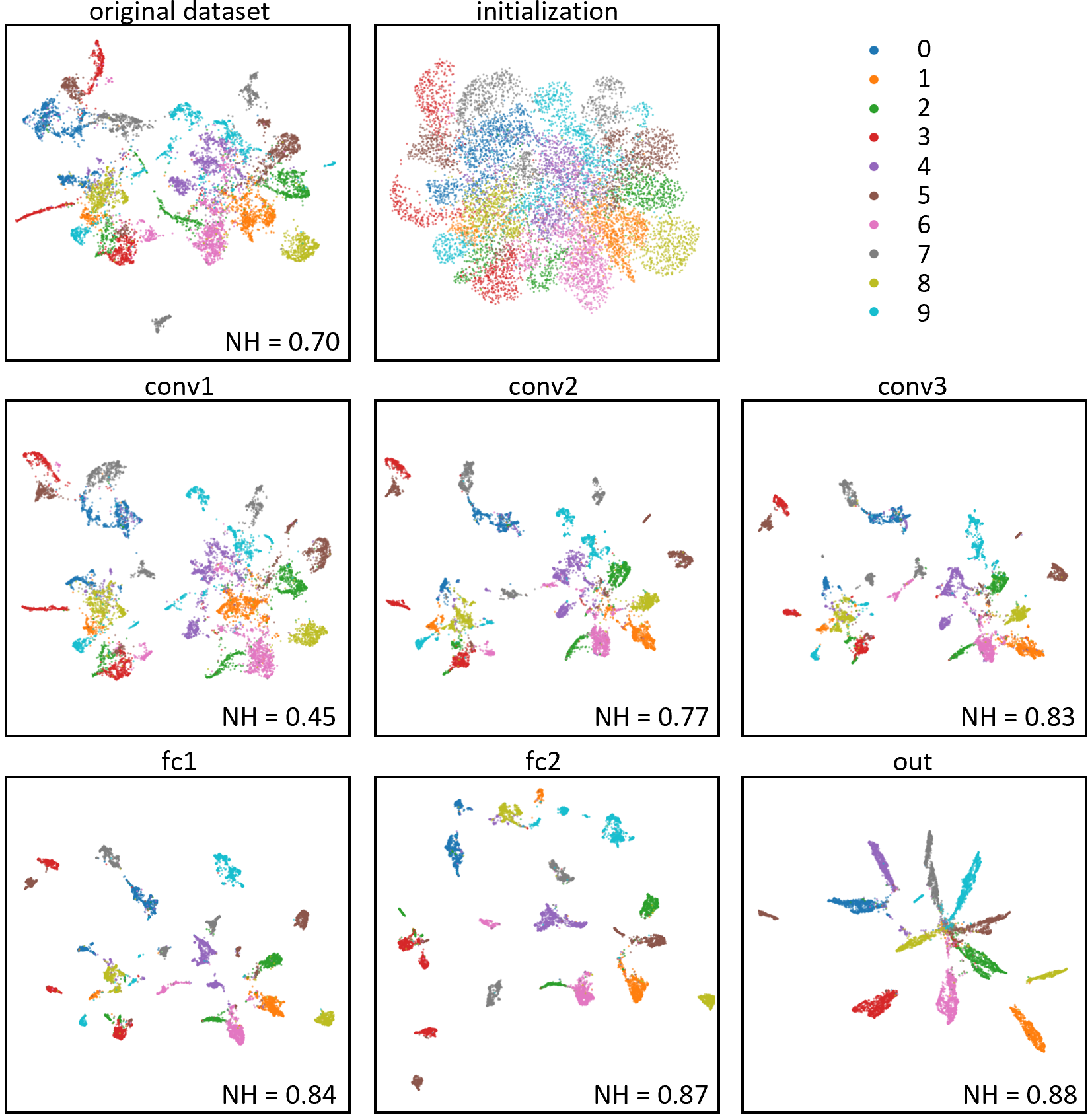}}
\caption{Evolvement of the learned representations along the layers of \textit{K-Net}. Data points are colored according to their target class. The top middle panel shows the initialization used for all layer embeddings.}
\label{fig:dev_knet_no_abl}
\end{center}
\vskip -0.2in
\end{figure}

We investigated how the learned representations evolve along the network layers and how they are affected by ablations. Figure \ref{fig:dev_knet_no_abl} shows the representations in the different network layers in the horizontally reduced activation space of the intact \textit{K-Net}.
We found that the separability of classes becomes clearer further down the network as indicated by the increasing NH-score, which is consistent with previous findings \cite{rauber2016visualizing}. In the representation in \textit{conv2} the NH-score is higher than the NH-score in the embedding of the original dataset, despite the higher dimensionality of the activation space compared to the original feature space, suggesting that at least two convolutional layers are necessary to extract meaningful features to separate the classes. In general, class clusters become more dense and more distinct from \textit{conv1} to \textit{fc1} and are bundled together after \textit{fc1} to \textit{out}. For example, comparing the representations in \textit{fc1} and \textit{fc2}, class 3 (red) and class 9 (cyan) are mapped closer together. There are exceptions to this trend however, e.g. for class 5 (brown) and class 8 (yellow), which remain split-up even after the soft-max activation in \textit{out}. This shows that the representation of \textit{out} is still able to represent more distinct classes than the number of labeled classes in the datasets.

\begin{figure*}[htb!]
\vskip 0.2in
\begin{center}
\centerline{\includegraphics[width=0.98\textwidth]{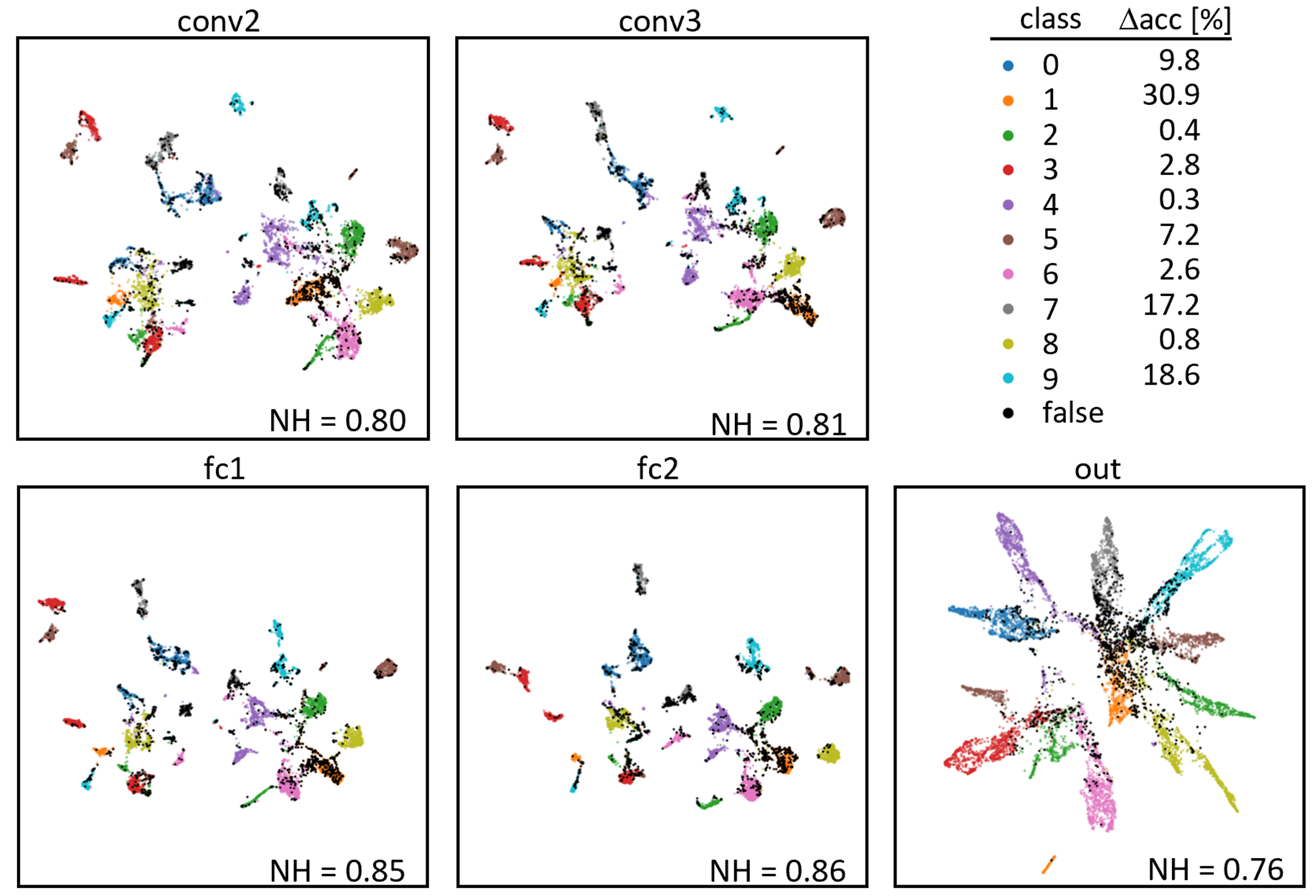}}
\caption{Effects of ablations in \textit{conv1} on the evolvement of the learned representations in subsequent layers in \textit{K-Net}. Black points represent the misclassified images as a result of the ablations.}
\label{fig:dev_knet_abl_full}
\end{center}
\vskip -0.2in
\end{figure*}

Subsequently, we investigated how the learned representations change after network ablations. We hypothesized, that ablations would locally distort the activation-space so that particularly heavily affected classes would be represented differently. Figure \ref{fig:dev_knet_abl_full} shows the layer representations of the ablated \textit{K-Net}. 
We performed ablations in \textit{conv1} and observed the strongest accuracy change of $30.9\%$p for class 1, and the weakest of $0.4\%$p for class 2. The black dots represent all the images that are misclassified as a result of the network ablations. Ablations had no particular effect on the separability of classes in the layer representations, as indicated by the NH-scores except in the output layer, where the misclassified images are not distributed into separate clusters anymore. Most of the black points correspond to misclassified images of class 1, which showed the highest drop in accuracy of $30.9\%$p. This suggests that the ablations caused a distortion in the representations of class 1, which amplified along the network layers and led to a less distinguishable representation of class 1 in the output layer. Interestingly, the representation of class 2 was not unified into a single cluster, as was the case for the healthy \textit{K-Net}. Considering the high inter-class variance of the KMNIST dataset, this implies that ablations deprived the network of representing some kind of similarly that would aggregate the different characters of class 2. Yet, the network did not lose much of the prediction performance for this class ($0.4\%$p). The Figures A3-A6 in the Appendix show similar effects for \textit{M-Net} and \textit{F-Net}.

\subsection{Functional Neuron Populations}
\begin{figure*}[htb!]
\vskip 0.2in
\begin{center}
\centerline{\includegraphics[width=0.98\textwidth]{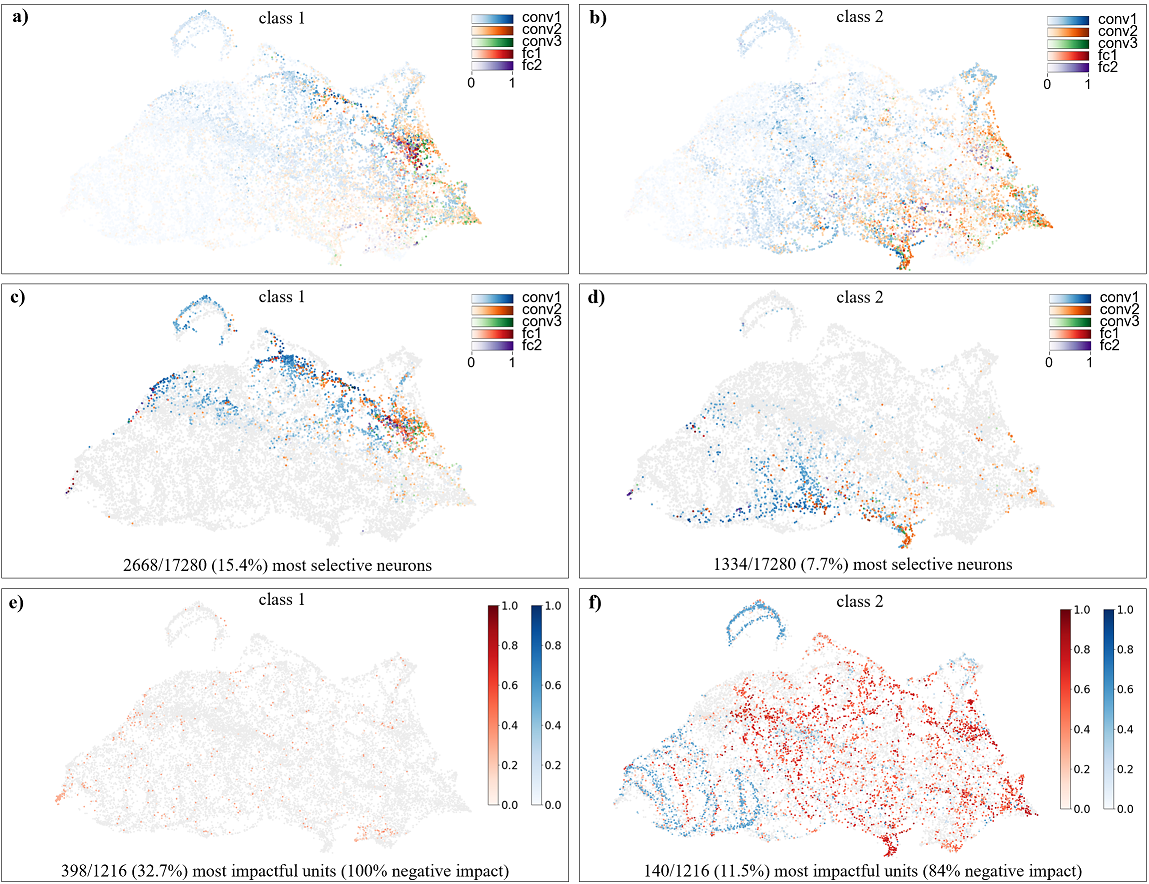}}
\caption{Neuron population of \textit{F-Net} with different color-codes. Neurons are colored according to layer affiliation and magnitude of activation (top row), according to layer affiliation and selectivity of their activation (middle row) and according to their impact on network accuracy upon ablation (bottom row). Left column: neuron activation was measured in response to a single example of class 1 (trouser) while the impact of ablations was calculated for all images of class 1 in the test set. Right column: same as left column, but for class 2.}
\label{fig:neuron_pop_collection}
\end{center}
\vskip -0.2in
\end{figure*}

Inspired by research in the field of neuroscience, we aimed to identify functional neuron populations in our networks, i.e. groups of neurons that a) show covariant behavior in response to input stimuli or b) affect the network accuracy in a similar way. Figure \ref{fig:neuron_pop_collection} shows the neuron population of \textit{F-Net} with different color-codes.

The neurons in Figure \ref{fig:neuron_pop_collection}a) and b) are colored according to their layer affiliations and their magnitude of activation in response to a single example image of class 1 and 2, respectively, where a strong/weak saturation corresponds to a strong/weak activation. The activations are normalized between values of $0$ and $1$ for each layer separately, due to large numerical differences of the absolute values across layers. Comparing both activation patterns with each other reveals that there are different clusters of neurons that jointly activate in response to the different stimuli, indicating that the different classes are represented by a different sets of neurons. 

\begin{figure*}[htb!]
\vskip 0.2in
\begin{center}
\centerline{\includegraphics[width=0.98\textwidth]{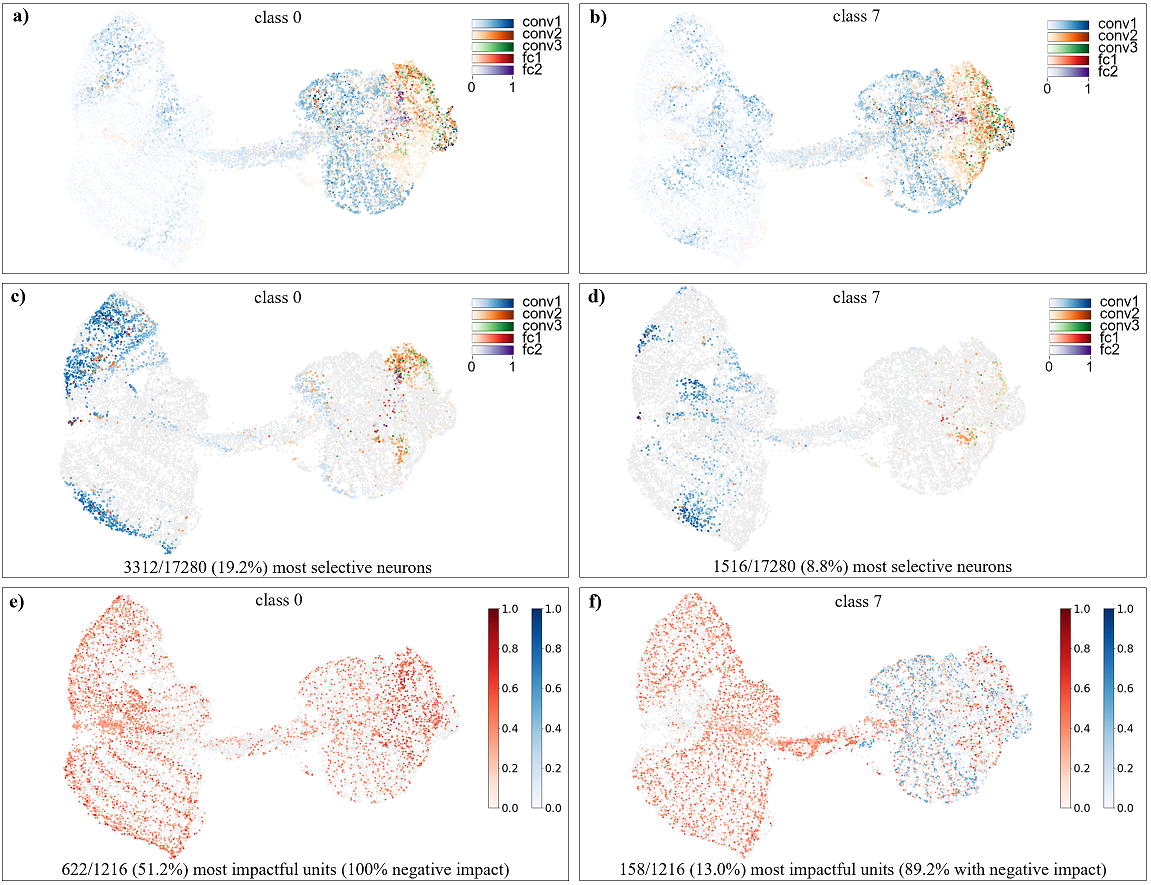}}
\caption{Neuron population of \textit{K-Net} with coloring analogous to Figure \ref{fig:neuron_pop_collection}.}
\label{fig:neuron_pop_collection_2}
\end{center}
\vskip -0.2in
\end{figure*}

The neurons in Figure \ref{fig:neuron_pop_collection}c) and d) are colored according to their AS. Colored neurons are most selective for class 1 and 2, respectively, while grey neurons are most selective for other classes. A strong/weak saturation in color corresponds to a high/low value of selectivity. Consistent to Figure \ref{fig:neuron_pop_collection}a) and b), the selectivity patterns reveal different clusters of most selective neurons for the different classes. Interestingly, these clusters differ from the clusters in \ref{fig:neuron_pop_collection}a) and b), suggesting that neurons, which are most selective for a specific class are not necessarily the most active in response to stimuli from that class. For both metrics, distinct clusters appear for all classes, but some regions are shared across classes. This implies that \textit{F-Net} represents both, class specific and cross class features in early layers. This is somewhat surprising, as early layers in CNNs are typically thought to represent general features shared by multiple classes \cite{Gatys2015ANA}. Furthermore, the amount of units that are most selective for a specific class varies greatly across the different classes. Specifically, more than twice as many neurons are most selective for class 1 compared to class 2. 

The neurons in Figure \ref{fig:neuron_pop_collection}e) and f) are colored according to their AES. Neurons, which negatively/positively impact the networks accuracy upon ablation are red/blue, while the saturation corresponds to the severity of the impact. A strong/weak saturation corresponds to a high/low impact, which is scaled between $0$ and $1$ (cf. section \ref{sec:embedding_of_net_act}). Note that for ablated units in the convolutional layers, 25 single neurons corresponding to the ablated kernel share the same AES value. Consistent with the previous results, the number of neurons that are most selective in their impact on network accuracy differs greatly across the different classes. Specifically, almost three times as many neurons are most selectively impacting the accuracy of class 1 compared to class two. However, all of those units show a negative impact, while a couple of units show a positive impact on network accuracy of class 2 upon ablation. This finding is consistent with previously reported results \cite{meyesablation} and suggests that ablations may be used to fine-tune network structure to improve network performance beyond its initial training accuracies. Interestingly, similar classes, i.e. classes that are close to each other in the UMAP embedding of the data, did not necessarily invoke similar patterns of activation, selectivity or ablation impact. We found that classes seem to arbitrarily share strongly activated areas or show exclusive patterns (cf. Figure \ref{fig:neuron_pop_collection_2} a) and b)). Figures S7 and S8 show similar findings for the neuron populations of \textit{M-Net} and \textit{K-Net}, even though for the latter, samples of the same class are largely different from each other.

We assessed, if classes with a high number of selective neurons are easier to separate than classes with a low number of selective neurons by calculating the Pearson correlation between the number of selective neurons per class and their NH-score calculated in the UMAP embeddings of the test dataset. A correlation coefficient of $r=0.61$ and a p-value of $p=0.06$ confirms a positive correlation, implying that classes that are well separable due to a higher number of class specific features are expectedly represented by a larger amount of selective neurons in the network. This positive correlation was only found for \textit{F-Net}, but not for \textit{M-Net} ($r=0.28$, $p=0.44$) or \textit{K-Net} (($r=0.30$, $p=0.40$)). Additionally, \textit{F-Net} showed a significant positive correlation between the number of units with the most class specific impact on network accuracy upon ablation and their NH-score ($r=0.68$, $p=0.03$). Again, such significance was not found for \textit{M-Net} ($r=-0.31$, $p=0.38$) or \textit{K-Net} ($r=0.50$, $p=0.14$). Note that in all cases the number of samples of $10$ is small, limiting the descriptive statistical power of the test.

\section{Conclusion and Outlook}
In this paper, we have taken an empirical approach to characterize the learned representations in neural networks to identify structural key elements aiming to describe their role for the task of the network. We found that the class specific representations are not evenly distributed across the network but localized either in specific layers or groups of neurons and that these distributions vary greatly across classes. This implies that the extent of the localization of knowledge depends on class specific properties and raises two questions. 1) How is the localization of a class specific representation affected by these class specific properties? 2) How does the robustness of these class specific representations against network ablations depend on these properties?

We further found that the learned representations evolve along the layers of the network to become more distinct facilitating better class separability in the network's activation-space. Network ablations in earlier layers only marginally affect the separability in subsequent layers but show a strong effect in the output layer. This suggests that ablations do not selectively affect parts of the network but rather the whole network in a holistic manner as the relative positions of single units in the activation space is mostly preserved. However, the distortion of the representation in the output layer implies that strongly class distinguishing features are still represented but more subtle features are not. This may be due to a redundant representation of such strongly class distinguishing features making their representation more robust against ablations than other features. Further work will be necessary to determine how such robustness can be characterized and achieved purposefully.

The finding of functional neuron populations revealed that the size of such populations differs greatly depending on their role to represent a specific class, raising the question how the required capacity of a network depends on the properties of that specific class. Furthermore, the lack of similarity between the functional neuron populations colored according to different metrics suggests that there is no single metric that sufficiently describes the role of single units within the whole network. In this context, we only investigated the effect of single unit ablations on network performance, however, this does not allow to determine whether the ablated unit is single handedly important or whether this unit is part of an important path through the network that has been altered by the ablation. We plan to address this issue in future work, aiming to identify such important paths along the network.

Concludingly, we argue that answering the question of how knowledge is represented in artificial networks is beneficial in two ways. First, a deeper understanding of how knowledge is represented and where it is localized in neural networks would facilitate the transfer of such knowledge from one system to another. Such insights would encourage new methods of transfer learning beyond the mere reuse of networks as feature extractors towards a modular recombination of important network paths and structures. Second, it addresses the issue of reproducibility in neuroscience, which despite modern experimental methods is one of the most critical issues stemming from the large differences between brains and the commonly small sample sizes in neuroscientific studies. Uncovering parallels between the structure and organization of represented knowledge in artificial and biological systems could be exploited and would provide measures and possibilities for initial large scale studies of artificial systems before transferring them to biological systems.
\small
\bibliographystyle{apacite}
\bibliography{main}

\newpage
\appendix
\onecolumn
\renewcommand\thefigure{\thesection.\arabic{figure}}    
\section{Appendix}
\setcounter{figure}{0}

\begin{figure}[b!]
\vskip 0.2in
\begin{center}
\centerline{\includegraphics[width=0.8\textwidth]{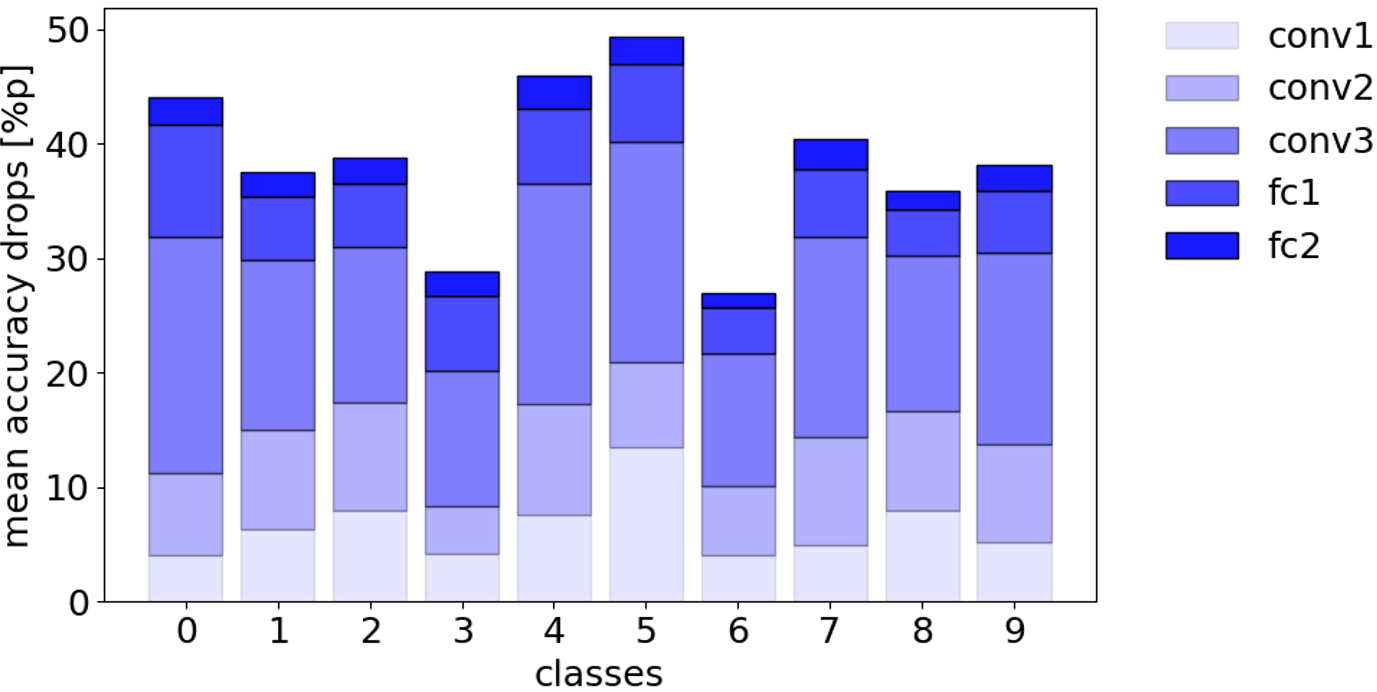}}
\caption{Stacked bar plot of mean accuracy changes as a result of network ablations in \textit{K-Net}.}
\end{center}
\vskip -0.2in
\end{figure}

\begin{figure}[b!]
\vskip 0.2in
\begin{center}
\centerline{\includegraphics[width=0.8\textwidth]{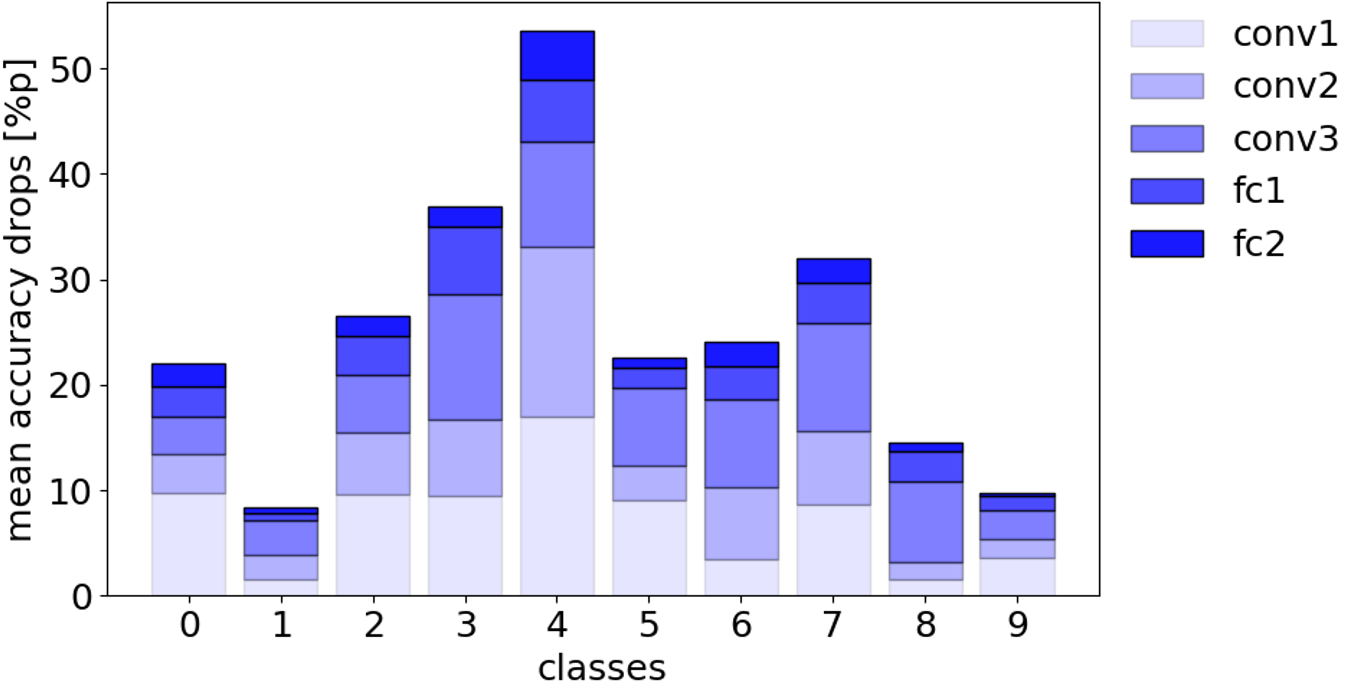}}
\caption{Stacked bar plot of mean accuracy changes as a result of network ablations in \textit{F-Net}.}
\end{center}
\vskip -0.2in
\end{figure}

\begin{figure}[b!]
    \centering
      \vspace*{0pt}
        \includegraphics[width=0.9\textwidth]{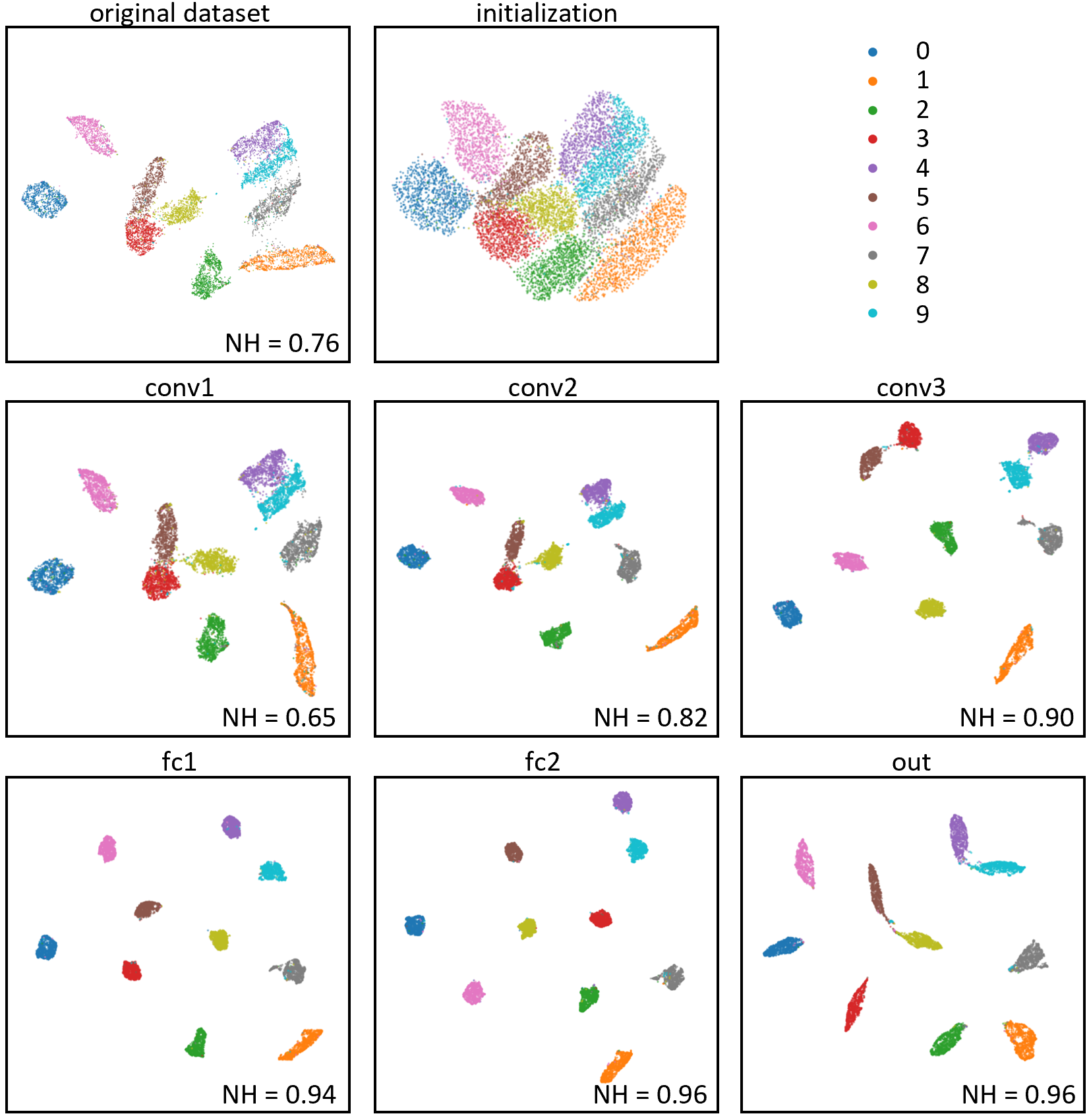}
        \caption{Change in representation, as data passes through \textit{M-Net}}
    \vspace*{-6pt}
\end{figure}

\begin{figure}[htb!]
    \centering
      \vspace*{0pt}
        \includegraphics[width=0.9\textwidth]{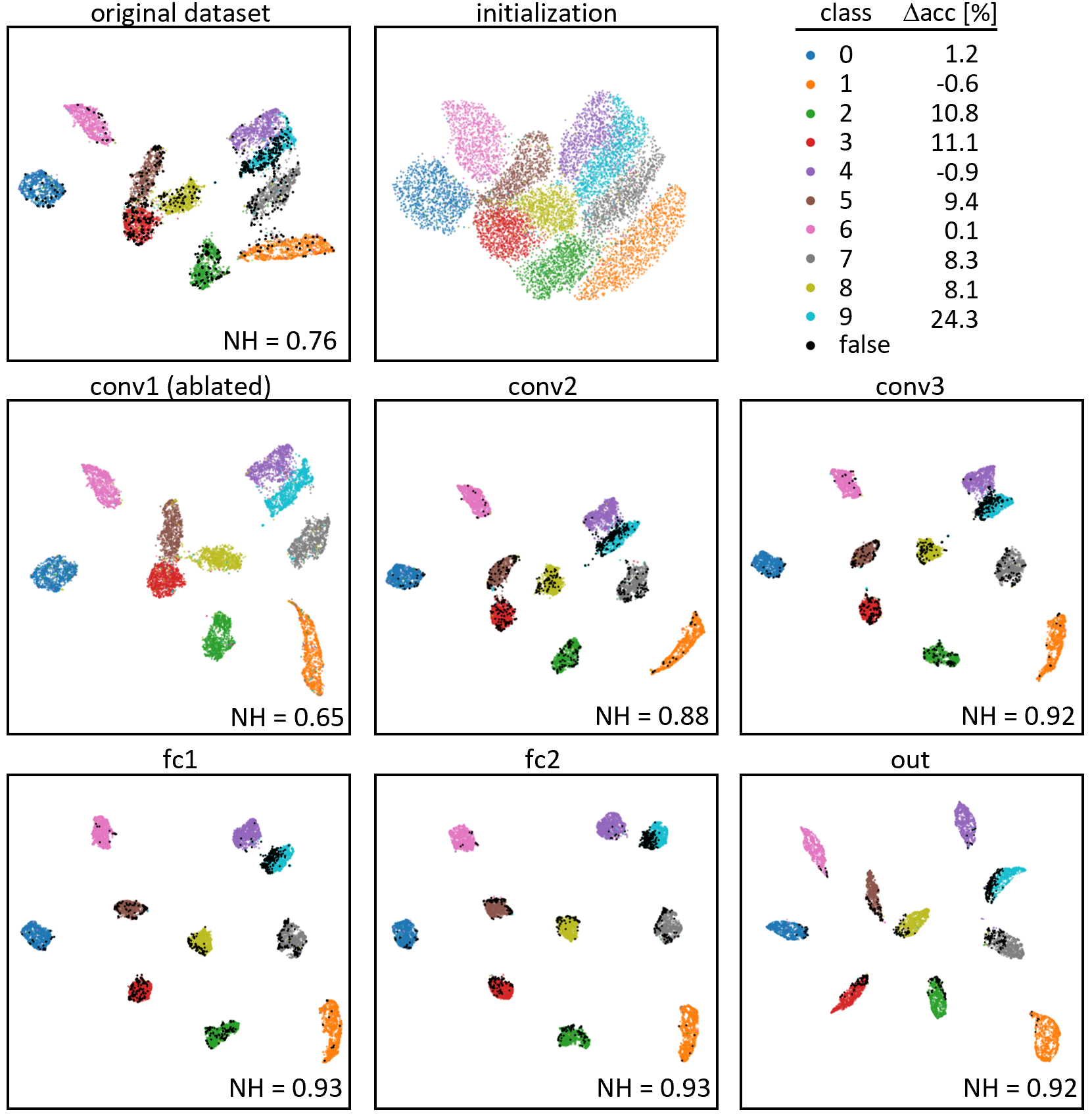}
        \caption{Change in representation, as data passes through an ablated version of \textit{M-Net}. The ablation was conducted with a random seed of $4$}
    \vspace*{-6pt}
\end{figure}

\begin{figure}[htb!]
    \centering
      \vspace*{0pt}
        \includegraphics[width=0.9\textwidth]{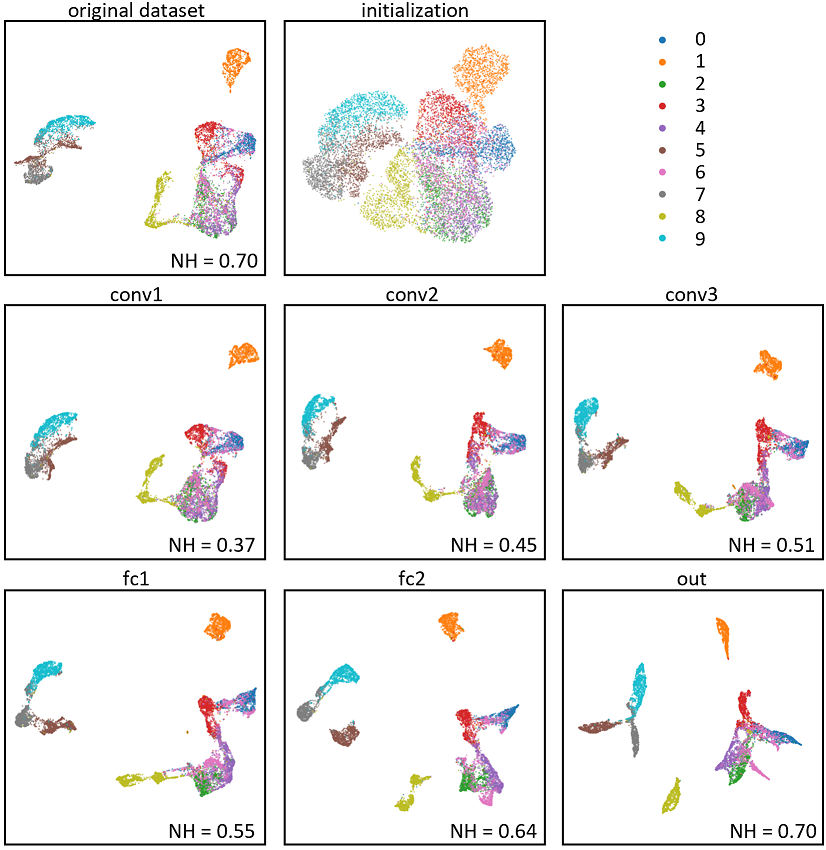}
        \caption{Change in representation, as data passes through \textit{F-Net}}
    \vspace*{-6pt}
\end{figure}

\begin{figure}[htb!]
    \centering
      \vspace*{0pt}
        \includegraphics[width=0.9\textwidth]{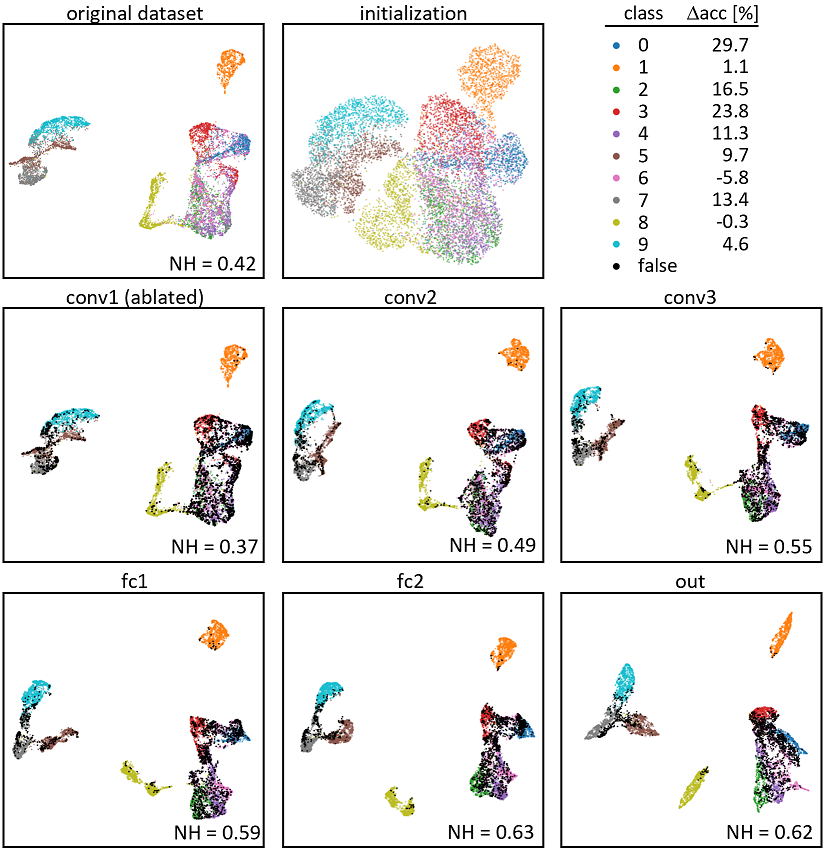}
        \caption{Change in representation, as data passes through an ablated version of \textit{
        F-Net}. The ablation was conducted with a random seed of $7$}
    \vspace*{-6pt}
\end{figure}

\begin{figure}[htb!]
    \centering
      \vspace*{0pt}
        \includegraphics[width=\textwidth]{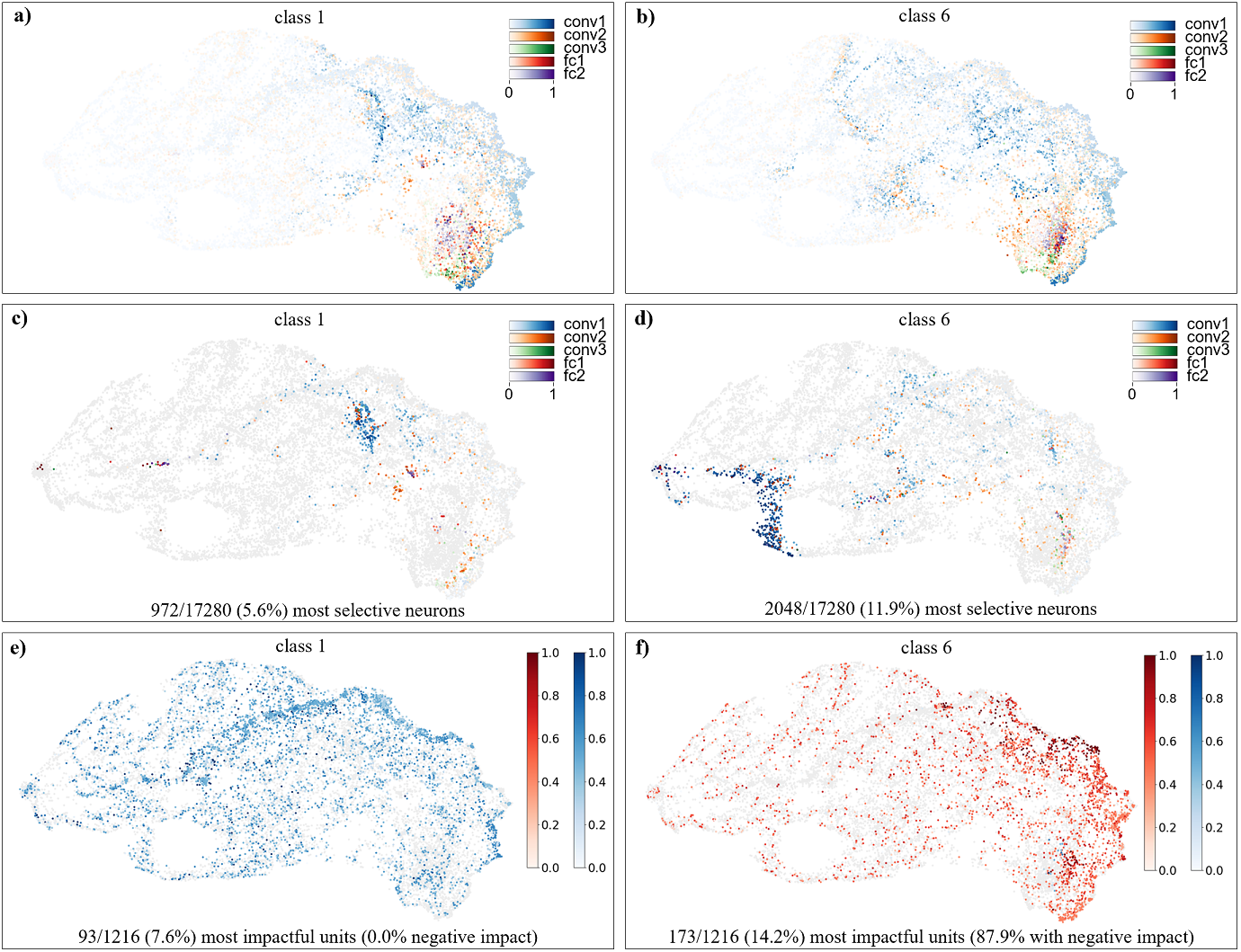}
        \caption{Neuron population of \textit{M-Net} color-coded by activational response to a single class sample, AS and AES for classes $1$ and $6$.}
    \vspace*{-6pt}
\end{figure}

\begin{figure}[htb!]
    \centering
      \vspace*{0pt}
        \includegraphics[width=\textwidth]{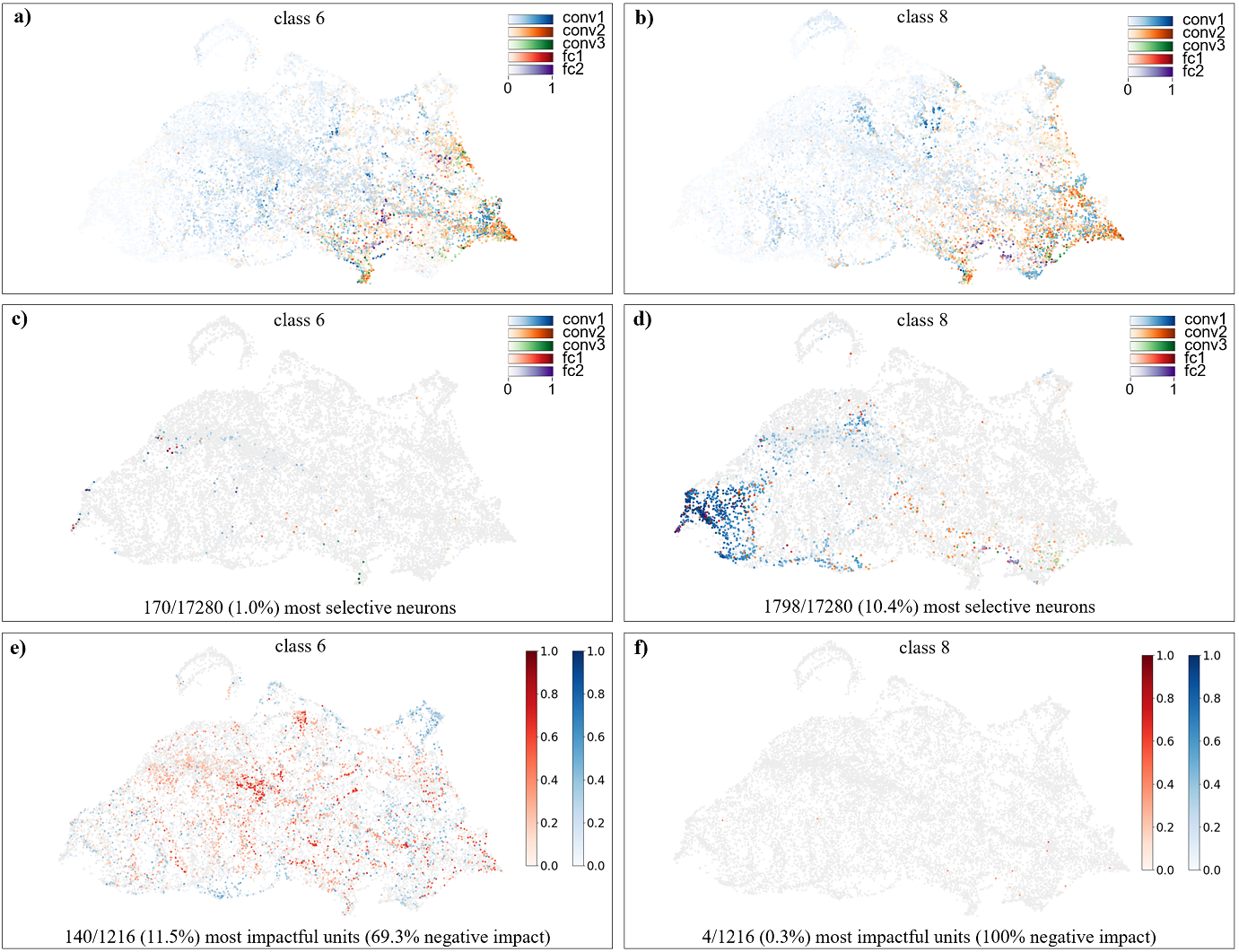}
        \caption{Neuron population of \textit{M-Net} color-coded by activational response to a single class sample, AS and AES for classes $0$ and $7$.}
    \vspace*{-6pt}
\end{figure}

\end{document}